\documentclass[conference]{IEEEtran}
\def\Ref#1{(\ref{#1})}

\newcommand{\diag}{{\rm diag}}

\newcommand{\nn}{\nonumber}
\newcommand{\al}[1]{\begin{eqnarray}#1\end{eqnarray}}
\newcommand{\eq}[1]{\begin{equation}#1\end{equation}}

\newcommand{\alequal}{&\!\!\!=\!\!\!&}

\ifCLASSINFOpdf
  \usepackage[pdftex]{graphicx}
\else
\fi
\usepackage[cmex10]{amsmath}
\begin{document}
%
\title{Observable dynamics and coordinate systems for automotive target tracking}

\author{\IEEEauthorblockN{Richard Altendorfer}
\IEEEauthorblockA{Driver Assistance Systems\\
TRW Automotive\\
Email: richard.altendorfer@trw.com}}



\maketitle

\begin{abstract}
We investigate several coordinate systems and dynamical vector
fields for target tracking to be used in driver assistance systems. We show
how to express the discrete dynamics of maneuvering target vehicles in
arbitrary coordinates starting from the target's and the own (ego) vehicle's
assumed dynamical model in
global coordinates. We clarify the notion of ``ego compensation'' and show
how non-inertial effects are to be included when using a body-fixed
coordinate system for target tracking. We finally compare the tracking error of
different combinations of target tracking coordinates and dynamical
vector fields for simulated data.
\end{abstract}


%
\IEEEpeerreviewmaketitle

\section{INTRODUCTION}

Driver assistance systems (DAS) such as adaptive cruise
control (ACC) or lane departure warning (LDW) need to perceive the environment using
exteroceptive sensors (e. g. radar or laser for ACC and camera
for LDW). As DAS become more sophisticated and move
from comfort applications to safety-critical applications such as automatic 
emergency braking, the requirements
regarding perception are becoming more stringent.
An essential part of exteroception is the target dynamics used in the
temporal prediction of observers with a predictor-corrector structure. 
In this paper different target dynamics in different
coordinate systems will be derived and assessed with respect to satisfactory modeling of target dynamics
and observability.

The modeling of target dynamics without an explicit estimation of the target driver's intent for DAS
is generally approximated by
assuming non-maneuver or maneuver models \cite{LiJilkov03,BarShalomKirubarajan01} of varying degrees of complexity \cite{Schubert_et_al_08}
where the target driver's actions like accelerating or steering are subsumed into (Gaussian) noise processes:
\eq{
\dot\xi^t_g = f^t(\xi^t_g, u^t_g, \nu^t_g) \nn
}
with $\xi^t_g$ the target state vector in global coordinates, $u^t_g$ the input (control) vector, and $\nu^t_g$ a multi-dimensional stochastic process.
Those models are usually formulated with respect to a global, inertial reference frame\footnote{The effect of the earth's rotation
around its polar axis as well as its rotation around the sun, etc on the vehicle motion can safely be neglected, hence an earth fixed reference
frame will be called ``inertial" in this paper.}
tangential to the 
earth's surface. This also applies to the modeling of the ownship (``ego") dynamics, 
$
\dot\xi^{ego}_g = f^{ego}(\xi^{ego}_g, u^{ego}_g, \nu^{ego}_g) 
$
however the corresponding observer can be fed
with proprioceptive measurements such
as yaw rate, lateral acceleration, or longitudinal velocity coming from vehicle stability control (VSC) systems.
Since in both target and ego modeling the input is usually zero, we will suppress the input vectors $u$ in these vector fields from now on.

Estimation of the target dynamics is based on exteroception by radar, laser, or video sensors
that provide measurements $\zeta$ relative to the ego vehicle. Hence somewhere in the estimation process a transformation from
relative to global coordinates must be performed.\footnote{Modeling the target dynamics {\it a priori} in coordinates relative to the ego
vehicle $\dot\xi^t_{rel} = f(\xi^t_{rel}, \nu^t_{rel})$ is rather unattractive and not considered here since e. g.
a constant acceleration model for $f$ would imply that the target permanently moves with a constant acceleration
plus noise relative to the ego vehicle irrespective of the actual state and motion of the ego vehicle.} A dynamical system for the 
combined target and ego observer would then
have the following form
\al{
\dot\xi^t_g \alequal f^t( \xi^t_g, \nu^t_g ) \nn\\
\dot\xi^{ego}_g \alequal f^{ego}(\xi^{ego}_g, \nu^{ego}_g) \nn\\
      \zeta^t \alequal h^t( \xi^t_g, u^t_g, w^t ) \nn\\
      \zeta^{ego} \alequal h^{ego}( \xi^{ego}_g, w^{ego} )\label{eq_target_dyn_global}
}
where the $w$'s are stochastic measurement processes.
Since the measurements $\zeta^t$ are relative to the ego vehicle, the output function $h^t$ must contain a control vector
$u^t_g = \xi^{ego}_g$ in order to map the relative measurements onto the global target state $\xi^t_g$. As all relative measurements
contain some sort of position information, the ego state $\xi^{ego}_g$ must hence also estimate the position of the ego vehicle.

This would be an appropriate approach if the ego state $\xi^{ego}_g$ were fully
observable which implies absolute position measurements by e. g. GPS receivers.
If $\xi^{ego}_g$ were not fully observable its covariance would grow without bounds which would cause 
the covariance matrix of $\xi^t_g$ to grow infinitely by propagation through the output function $h^t$. 
This means that the dynamical system not stochastically observable -- a necessary condition for the convergence of an
extended Kalman filter (EKF), see e. g. \cite{SongGrizzle95}.  
Since most vehicles are not equipped with a GPS receiver, their
position can only be estimated by dead-reckoning and is therefore unobservable. Even if vehicles are equipped with a
GPS receiver e. g. from their navigation system, the GPS position (without differential corrections)
is only accurate to about $\pm 5\dots 15m$ \cite{Wikipedia_Global_Positioning_System}.
This error would then be propagated to unacceptably large position covariances in $\xi^t_g$.
While the covariance of the {\it relative} position in this case might remain bounded as suggested in 
 \cite{BuehrenYang07} it is not advisable to work with unobservable, non-convergent systems whose
ever-growing covariances will also invalidate the propagation of covariance by linearization as in an EKF. 

Hence we propose formulating the target dynamics in {\it relative}, i. e. in ego vehicle fixed coordinates.
The goal is to replace the combined dynamical system \Ref{eq_target_dyn_global} with a system that contains $\xi^t_{rel}$ and $\xi^{ego}_g$ 
where the unobservable states of $\xi^{ego}_g$ are not used for the estimation of $\xi^t_{rel}$.
In the next section it will be shown how to derive the relative target dynamics starting from the
global target and ego dynamics in eq. \ref{eq_target_dyn_global}. 
While the use of relative target dynamics for automotive target tracking is not new (see e. g. \cite{MoebusJoosKolbe03,Stueker_04,Maehlisch_et_al_07}), in this paper a general, system-theoretic framework for the rigorous
derivation of relative target dynamics starting from arbitrary dynamical vector fields for global target and ego dynamics is presented.
This includes the derivation of the discrete dynamics with process and input noise covariance matrices as needed for an 
EKF as observer. The derivation is illustrated by three different choices of coordinate systems and vector fields and their accuracy in tracking targets is
assessed by numerical simulation. The observability of the combined target and ego 
dynamics is also discussed.

\begin{figure}[t]
\centering
\includegraphics[width=1.8in]{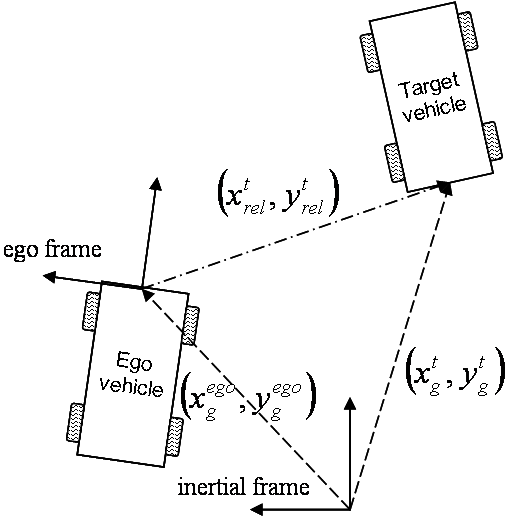}
\caption{Inertial and ego-fixed coordinate frames. The origin of the ego-fixed frame and the reference point of
the target vehicle are common choices but are by no means unique.}
\label{fig_coordinate_systems}
\end{figure}

\section{RELATIVE TARGET VEHICLE DYNAMICS}\label{sec_relative_vehicle_dynamics}

\subsection{Derivation}

In order to derive the discrete target dynamics in relative coordinates we first need to obtain its continuous vector field.
This requires a definition of the state vectors for the global target and ego dynamics and their continuous vector fields
as well as a definition of the relative coordinates.
The vector fields are given by
\al{
\dot\xi^{ego}_g \alequal f^{ego}(\xi^{ego}_g, \nu^{ego}_g) \nn\\
    \dot\xi^t_g \alequal f^{t}(\xi^t_g, \nu^t_g) \label{eq_target_vector_field}
}
where we assume that no control inputs are necessary for $f^{ego}$ and $f^{t}$.
We now define a new state by
\eq{
\xi^t_{rel} = m( \xi^t_g, \xi^{ego}_g ) \label{eq_general_relative_state}
}
where $m$ is in general a nonlinear function that depends upon the choice of
the coordinates of $\xi^t_g$, $\xi^{ego}_g$ (Cartesian, polar, etc). If both $\xi^t_g$ and $\xi^{ego}_g$
are in Cartesian coordinates and the new coordinates are also Cartesian coordinates for a ego body fixed
system we get the more intuitive expression
\eq{
m( \xi^t_g, \xi^{ego}_g ) = M(\xi^{ego}_g)\left( \xi^t_g - \xi^{ego}_g \right) \label{eq_definition_coordinates}
}
where $M(\xi^{ego}_g)$ contains a rotation to the ego-fixed coordinate system as well as corrections due to the fact that
the ego-fixed system is a non-inertial system and thus experiences pseudo-forces, see app. \ref{app_rotations}.\footnote{From now on, we suppress
the dependence of $M$ on $\xi^{ego}_g$ in our notation.}
By taking the time derivative of eq. \Ref{eq_definition_coordinates}
we get a vector field for $\xi^t_{rel}$
\eq{
\dot\xi^t_{rel} = \dot M ( \xi^t_g - \xi^{ego}_g ) + M \left( f^{t}(\xi^t_g, \nu^t_g) - f^{ego}(\xi^{ego}_g, \nu^{ego}_g) \right)\nn
}


However, the goal of this computation is to replace $\xi^t_g$ with $\xi^t_{rel}$; therefore we need to use
eq. \Ref{eq_definition_coordinates} again in order to eliminate $\xi^t_g$:
\al{
\dot\xi^t_{rel} \alequal \dot M M^{-1}\xi^t_{rel}\label{eq_cont_dynamics}\\
& &\! + M\left(f^{t}(M^{-1}\xi^t_{rel}+\xi^{ego}_g, \nu^t_g) - f^{ego}(\xi^{ego}_g, \nu^{ego}_g) \right)\nn
}
The combined system of differential equations for $\xi^t_{rel}$ and $\xi^{ego}_g$ reads
\al{
\dot\xi^t_{rel} \alequal \dot M M^{-1}\xi^t_{rel}\nn\\
& & + M\left(f^{t}(M^{-1}\xi^t_{rel}+\xi^{ego}_g, \nu^t_g) - f^{ego}(\xi^{ego}_g, \nu^{ego}_g) \right)\nn\\
\dot\xi^{ego}_g \alequal f^{ego}(\xi^{ego}_g, \nu^{ego}_g)\label{eq_combined_cont_dynamics}
}
This system of differential equations governs the dynamics of the target vehicle relative to the ego vehicle and are based upon:
the dynamics of the target vehicle with respect to the ground $f^t$, the dynamics of the ego vehicle with respect to the ground $f^{ego}$, and
the definition of the relative coordinates $m$. 
This procedure separates dynamical models for individual vehicle dynamics (ego or target)
from the relative dynamics used for tracking in an arbitrary coordinate system.
The ``ego compensation" at the level of continuous dynamics is not a separate step
but is intertwined with the relative dynamics.

\subsection{Solution}

In this paper, all stochastic differential equations are chosen to be solved by the 
discrete-time counterpart method \cite{LiJilkov03} where the continuous stochastic process $\nu^{}_g(t)$ is 
replaced by a discrete stochastic process $\nu^{}_{g,k}$ which is constant 
from one time step to the next.\footnote{Alternatively, the discrete-time equivalent method \cite{LiJilkov03} can
be employed using the power spectral density of the continuous stochastic process.}
By abuse of notation, here $\nu^{}_g$ also denotes this constant value.

Since the ego dynamics is decoupled from the relative target dynamics, it can be solved first, see 
app. \ref{app_CTRA_model}, and its 
solution $\xi^{ego}_g(t) = F^{ego}_g( \xi^{ego}_g(t_0), t-t_0, \nu^{ego}_g )$
can be inserted into the differential equation for $\xi^t_{rel}$ to arrive at a time-dependent differential equation:
\al{
\dot\xi^t_{rel} \alequal \dot M M^{-1}\xi^t_{rel}\nn\\
& & + M\left(f^{t}(M^{-1}\xi^t_{rel}+\xi^{ego}_g(t), \nu^t_g) - \dot\xi^{ego}_g(t)\right)\nn\\
\alequal f_{rel}(\xi^t_{rel}, \xi^{ego}_g(t), \dot\xi^{ego}_g(t), \nu^t_g, \nu^{ego}_g ) \label{eq_cont_dynamics_general}
}
The solution of this differential equation - if it exists - can be cast into the notation of discrete time systems to be used later
for application of the EKF:
\al{
\xi^{ego}_{g\ k+1} \alequal F^{ego}_g( \xi^{ego}_{g\ k}, \Delta t_k, \nu^{ego}_{g\ k} ) \nn\\
\xi^t_{rel\ k+1} \alequal F^{t}_{rel}( \xi^t_{rel\ k}, \xi^{ego}_{g\ k}, \Delta t_k, \nu^{t}_{rel\ k} ) \label{eq_discrete_time_solution}
}
where $\Delta t_k = t_{k+1} - t_k$ is the time difference from one iteration to the next and 
$\nu^{t}_{rel\ k} = ( \nu^t_{g\ k},\ \nu^{ego}_{g\ k} )^\top$ is the effective discrete stochastic process for the
relative target dynamics. Note that $\xi^{ego}_{g\ k+1}$ does not have inputs but $\xi^t_{rel\ k+1}$ has the
ego state $\xi^{ego}_{g\ k}$ as its input or control vector. 

In the following we will assume that the ego estimation using proprioceptive measurements from VSC sensors is
separate from the exteroception and only outputs the ego state $\xi^{ego}_{g\ k}$ and its covariance matrix $P^{ego}_{g\ k}$.

For the use of eq. \Ref{eq_discrete_time_solution} in an EKF, we define the matrices
\al{
A_k \alequal\partial_{\xi^t_{rel}} F^{t}_{rel}( \xi^t_{rel\ k}, \xi^{ego}_{g\ k}, \Delta t_k, \nu^{t}_{rel\ k} ) \nn\\
B_k \alequal\partial_{\xi^{ego}_{g}} F^{t}_{rel}( \xi^t_{rel\ k}, \xi^{ego}_{g\ k}, \Delta t_k, \nu^{t}_{rel\ k} ) \nn\\
G_k \alequal\partial_{\nu^{t}_{rel}} F^{t}_{rel}( \xi^t_{rel\ k}, \xi^{ego}_{g\ k}, \Delta t_k, \nu^{t}_{rel\ k} ) \nn
}
which are used for the propagation of the state ($A_k$), input ($B_k$), and process noise ($G_k$) 
covariances by linearization. The input noise and process noise covariance matrices are then
\al{
Q^{input}_k \alequal B_k P^{ego}_{g\ k} B_k^\top\nn\\
Q^{process}_k \alequal G_k V^{t}_{rel\ k} G_k^\top\nn
}
where $V^{t}_{rel\ k} = {\rm cov}( \nu^{t}_{rel\ k} )$.
Since nonlinear effects in estimation are not the focus of this paper we content ourselves in the numerical analysis with the standard 
EKF\footnote{The local convergence of the EKF for nonlinear discrete-time systems has been proven in \cite{SongGrizzle95} under certain conditions.}
unlike in \cite{Schubert_et_al_08} where the unscented Kalman filter was used. Note however, that even the unscented
transformation does not fully propagate the second moments of a multi-dimensional probability distribution \cite{Hendeby_Gustafsson_08}.


\subsection{Vehicle dynamics}\label{sec_vehicle_dynamics}

In eq. \Ref{eq_combined_cont_dynamics} arbitrary vector fields $f^{t}$ and $f^{ego}$ can be used, in this paper, however, we 
focus on two common object dynamics, namely the white noise jerk model
(see e. g. \cite{LiJilkov03}) and the constant turn rate and acceleration model (CTRA, see app. \ref{app_CTRA_model}). 
While the former might be advantageous in the 
object initialization stage if velocity and/or acceleration are not directly measured and are initialized by
default values or if not just vehicles but also other objects such as pedestrians are perceived,
the latter better represents the non-holonomic behavior of wheeled vehicles 
and also performed best in a comparative study of vehicle motion models \cite{Schubert_et_al_08}.
For the ego dynamics only the CTRA model is considered in this paper.

\subsection{Relative coordinate systems}

After choosing the target and ego dynamics the relative coordinate system for target tracking $m$ must be
specified. We will investigate two different coordinate systems, one in which the entire target motion is relative
to the ego vehicle (pure relative coordinates) and one in which the target position is estimated relative to the ego vehicle whereas the velocity
and acceleration are estimated over ground, but rotated into the ego coordinate system (``mixed" coordinates). 
We will not dwell on other aspects of coordinate system choices such as the performance of polar versus
Cartesian coordinates for target tracking as in \cite{GustafssonIsaksson96}.

\section{Examples}\label{sec_examples}

\subsection{Tracking in relative coordinates, white noise jerk model}

For this example the global target dynamics is the non-maneuver white noise jerk model. It is 
naturally expressed in Cartesian coordinates,
see e. g. \cite{LiJilkov03}. The ego dynamics is given by the CTRA model. The relative dynamics is expressed
in Cartesian coordinates. In order to apply eq. \Ref{eq_general_relative_state}, the
ego state vector must be transformed to Cartesian coordinates as well. With this choice the
function $m$ reads
\eq{
\xi^t_{rel} = m( \xi^t_g, \xi^{ego}_g ) = M(\xi^{ego}_g)\left( \xi^t_g - \xi^{ego}_g \right) \nn
}
as in \Ref{eq_definition_coordinates} and we obtain as in \Ref{eq_cont_dynamics}
\al{
\dot\xi^t_{rel} \alequal \dot M M^{-1}\xi^t_{rel}\nn\\
& & + M\left(f^{t}(M^{-1}\xi^t_{rel}+\xi^{ego}_g, \nu^t_g) - f^{ego}(\xi^{ego}_g, \nu^{ego}_g)\right)\nn
}
Using the ego trajectory eq. \Ref{eq_solution_CTRA} and defining
$\xi^t_{rel} = \begin{pmatrix} x & y & \dot x & \dot y & \ddot x & \ddot y \end{pmatrix}^\top$
we obtain
\eq{
\dot\xi^t_{rel} = \begin{pmatrix} \dot x, & \dot y, & \ddot x, & \ddot y, & \dddot x, & \dddot y \end{pmatrix}^\top \nn
}
with
\al{
\dddot x \alequal-\nu^{ego}_{\dot a,0}+3 {\dot\psi^{ego}_0} {\ddot y}+{1\over 2} (\dot\psi^{ego}_0)^2 \nu^{ego}_{\dot a,0} t^2+(\dot\psi^{ego}_0)^2 {a^{ego}_0} t\nn\\
&&-(\dot\psi^{ego}_0)^3 y+(\dot\psi^{ego}_0)^2 {v^{ego}_0}+2 \nu^{ego}_{\ddot\psi,0} (\dot\psi^{ego}_0) t^3 \nu^{ego}_{\dot a,0}\nn\\
&&+3 \nu^{ego}_{\ddot\psi,0} {\dot\psi^{ego}_0} t^2 {a^{ego}_0}+2 \nu^{ego}_{\ddot\psi,0} {\dot\psi^{ego}_0} t {v^{ego}_0}+3 (\dot\psi^{ego}_0)^2 {\dot x}\nn\\
&&+\cos({\dot\psi^{ego}_0} t+{\psi^{ego}_0}) \nu^{t}_{\dddot y,0}+\sin({\dot\psi^{ego}_0} t+{\psi^{ego}_0}) \nu^{t}_{\dddot x,0}\nn\\
\dddot y \alequal{1\over 4} \nu^{ego}_{\ddot\psi,0} (\dot\psi^{ego}_0)^2 t^4 \nu^{ego}_{\dot a,0}+{1\over 2} \nu^{ego}_{\ddot\psi,0} (\dot\psi^{ego}_0)^2 t^3 {a^{ego}_0}\label{eq_pure_relative_highest_derivative}\\
&&+{1\over 2} \nu^{ego}_{\ddot\psi,0} (\dot\psi^{ego}_0)^2 t^2 {v^{ego}_0}+(\dot\psi^{ego}_0)^3 x+3 (\dot\psi^{ego}_0)^2 {\dot y}\nn\\
&&-2 {\dot\psi^{ego}_0} {a^{ego}_0}-\nu^{ego}_{\ddot\psi,0} {v^{ego}_0}-3 {\dot\psi^{ego}_0} {\ddot x}\nn\\
&&-2 {\dot\psi^{ego}_0} \nu^{ego}_{\dot a,0} t-3 \nu^{ego}_{\ddot\psi,0} \nu^{ego}_{\dot a,0} t^2-3 \nu^{ego}_{\ddot\psi,0} {a^{ego}_0} t\nn\\
&&+\cos({\dot\psi^{ego}_0} t+{\psi^{ego}_0}) \nu^{t}_{\dddot x,0}-\sin({\dot\psi^{ego}_0} t+{\psi^{ego}_0}) \nu^{t}_{\dddot y,0}\nn
}
Note that the non-trivial components of this differential equation are - as expected - the highest derivatives; everything else is purely kinematic.
This would not have been the case had we omitted the non-inertial contributions in $M$.

In eq. \Ref{eq_cont_dynamics_general} the target state vector $\xi^t_{rel}$ is rotated into the ego coordinate system.
This rotation, however, is not applied to the target process noise $\nu^t_g$. On the other hand, the relative target
dynamics should not depend on the orientation of an arbitrary global coordinate system. 
In \Ref{eq_pure_relative_highest_derivative} the terms proportional to $\nu^{t}_{\dddot x,0}$ and $\nu^{t}_{\dddot y,0}$ depend on the global orientation $\psi^{ego}_0$ of the ego vehicle.
It can be checked, however, that by choosing the process noise to be isotropic, i. e. 
with identical covariance values for $\nu^{t}_{\dddot x,0}$ and $\nu^{t}_{\dddot y,0}$, the process covariance matrix $Q^{process}_k$ becomes independent of $\psi^{ego}_0$.

This differential equation is of the form $\dot\xi = A\xi + B(t)$ and can therefore be solved by the standard formula for linear
time-invariant systems which is also valid for time-variant $B(t)$, see e. g. \cite{Bay99}.
Since the solution $\xi^t_{rel\ k+1}$ and the corresponding Jacobians $A_k$, $B_k$, and $G_k$ are rather
unwieldy expressions which can easily be computed by standard symbolic computation engines such as Matlab's symbolic toolbox,
we will not provide them here.

\subsection{Tracking in mixed coordinates, white noise jerk model}

For this example the global target dynamics is again the non-maneuver white noise jerk model. However, in ``mixed"
coordinates, velocities and accelerations are the inertial quantities measured over ground, 
rotated into the ego coordinate system. This has the advantage that the
dynamics of velocity and acceleration is reduced in the sense that the range of values of the velocity and
acceleration is cut in half. This is particularly important for the object initialization of not measured states like acceleration.

Again we transform the ego state vector to Cartesian coordinates: 
$\xi^{ego}_{g} = \begin{pmatrix} x^{ego} & y^{ego} & \dot x^{ego} & \dot y^{ego} & \ddot x^{ego} & \ddot y^{ego} \end{pmatrix}^\top$.
Then we use matrix $M$ without non-inertial terms since the velocities and accelerations
are now inertial quantities
and introduce a projector $\Pi$ to project out the ego velocities and accelerations
\al{
\xi^t_{mix} \alequal m( \xi^t_g, \xi^{ego}_g ) = R(\xi^{ego}_g)\left( \xi^t_g - \Pi\xi^{ego}_g \right) \nn\\
\Pi \alequal {\rm diag}( 1, 1, 0, 0, 0, 0 ) \nn\\
R \alequal \begin{pmatrix} r & 0 & 0 \cr
                      0 & r & 0 \cr
                      0 & 0 & r \end{pmatrix} \nn  
}
The resulting differential equation reads
\al{
\dot\xi^t_{mix}\alequal\dot R R^{-1}\xi^t_{mix}\label{eq_mixed_wnjm}\\
&&\!\!\!\!\!\!\!\!\!\!\! + R\left(f^{t}(R^{-1}\xi^t_{mix}+\Pi\xi^{ego}_g, \nu^t_g) - \Pi f^{ego}(\xi^{ego}_g, \nu^{ego}_g)\right)\nn
}
Using the ego trajectory eq. \Ref{eq_solution_CTRA} and redefining
$\xi^t_{mix} = \begin{pmatrix} x & y & \dot x & \dot y & \ddot x & \ddot y \end{pmatrix}^\top$
we obtain
\eq{
\dot\xi^t_{rel} = \label{eq_mixed_white_noise_jerk_vector_field}
}
$$
\begin{pmatrix} 
{\dot\psi^{ego}_0} y+{\dot x}-{v^{ego}_0}-{1\over 2} \nu^{ego}_{\dot a,0} t^2-{a^{ego}_0} t \cr
-{\dot\psi^{ego}_0} x+{\dot y}-{1\over 2} \nu^{ego}_{\ddot\psi,0} {a^{ego}_0} t^3-{1\over 2} \nu^{ego}_{\ddot\psi,0} {v^{ego}_0} t^2 \cr
{\dot y} {\dot\psi^{ego}_0}+{\ddot x} \cr
-{\dot x} {\dot\psi^{ego}_0}+{\ddot y} \cr
{\ddot y} {\dot\psi^{ego}_0}+\cos({\scriptstyle {\dot\psi^{ego}_0} t+{\psi^{ego}_0}}) \nu^{t}_{\dddot x,0}+\sin({\scriptstyle {\dot\psi^{ego}_0} t+{\psi^{ego}_0}}) \nu^{t}_{\dddot y,0} \cr
-{\ddot x} {\dot\psi^{ego}_0}-\sin({\scriptstyle {\dot\psi^{ego}_0} t+{\psi^{ego}_0}}) \nu^{t}_{\dddot x,0}+\cos({\scriptstyle {\dot\psi^{ego}_0} t+{\psi^{ego}_0}}) \nu^{t}_{\dddot y,0}
\end{pmatrix}
$$
As in \Ref{eq_pure_relative_highest_derivative}, the terms proportional to $\nu^{t}_{\dddot x,0}$ and
$\nu^{t}_{\dddot y,0}$ depend on the global orientation $\psi^{ego}_0$ of the ego vehicle.
By choosing the process noise to be isotropic 
the process covariance matrix again becomes independent of $\psi^{ego}_0$.

This differential equation is also of the form $\dot\xi = A\xi + B(t)$ and can be solved by linear system techniques.
Again we will not provide the unwieldy but easily computable solution.

\subsection{Tracking in mixed coordinates, CTRA model}

Using the CTRA model for the representation of the target dynamics, it is more convenient not to transform to Cartesian coordinates.
Hence both $\xi^{ego}_{g}$ and $\xi^{t}_{g}$ are interpreted as $\begin{pmatrix} x_{rel} & y_{rel} & \psi_{rel} & \dot\psi_{g} & v_{g} & a_{g} \end{pmatrix}^\top$.
The relative coordinates are the relative $x_{rel}$ and $y_{rel}$ position with respect to the ego vehicle coordinate system,
the angle between the ego and target velocities over ground $\psi_{rel}$ which coincides for the CTRA models with the
relative angle of the vehicle orientations, the target yaw rate over ground $\dot\psi_{g}$ and the target speed $v_g$ and longitudinal acceleration $a_g$ over ground.

The coordinate transformation for this setting is
\al{
\xi^t_{mix} \alequal m( \xi^t_g, \xi^{ego}_g ) = \tilde R(\xi^{ego}_g)\left( \xi^t_g - \Pi\xi^{ego}_g \right) \nn\\
{\rm with} && \nn\\
\Pi \alequal {\rm diag}( 1, 1, 1, 0, 0, 0 )  \nn\\
\tilde R \alequal \begin{pmatrix} r & 0  & 0 \cr
                       0 & id & 0 \cr
                       0 & 0  & id \end{pmatrix} \nn
}
The resulting differential equation reads
\al{
\dot\xi^t_{mix}\alequal\dot{\tilde R} {\tilde R}^{-1}\xi^t_{mix}\nn\\
&&\!\!\!\!\!\!\!\!\!\!\!  + \tilde R\left(f^{t}({\tilde R}^{-1}\xi^t_{mix}+\Pi\xi^{ego}_g, \nu^t_g) - \Pi f^{ego}(\xi^{ego}_g, \nu^{ego}_g)\right)\nn
}

Using the ego trajectory eq. \Ref{eq_solution_CTRA} and defining
$\xi^t_{mix} = \begin{pmatrix} x_{rel} & y_{rel} & \psi_{rel} & \dot\psi_g & v_g & a_g \end{pmatrix}^\top$
we obtain
\eq{
\dot\xi^t_{mix} = \label{eq_CTRA_mixed_vector_field}
}
$$
\begin{pmatrix} 
y_{rel} {\dot\psi^{ego}_0}+{v_g} \cos({\scriptstyle {\psi_{rel}}+\nu^{ego}_{\ddot\psi, 0} {t^2\over 2}})- \nu^{ego}_{\dot a, 0} {t^2\over 2}-{a^{ego}_0} t-{v^{ego}_0}\cr
-x_{rel} {\dot\psi^{ego}_0}+{v_g} \sin({\scriptstyle {\psi_{rel}}+\nu^{ego}_{\ddot\psi, 0} {t^2\over 2}})- \nu^{ego}_{\ddot\psi, 0} { t^3 {a^{ego}_0}+ t^2 v^{ego}_0\over 2}\cr
{\dot\psi_g}-\nu^{ego}_{\ddot\psi, 0} t-\dot\psi^{ego}_0\cr
\nu^{t}_{\ddot\psi, 0}\cr
{a_g}\cr
\nu^{t}_{\dot a, 0}
\end{pmatrix}
$$
This non-linear differential equation can be solved as follows: 
the lower four entries of \Ref{eq_CTRA_mixed_vector_field} can easily be solved by direct integration.
Those expressions can now be inserted into the first two components. They are then of the form $\dot\xi = A\xi + B(t)$ and can be solved by
linear system techniques. Since terms proportional to $t^2$ appear inside the sine and cosine, integration
results in Fresnel integrals. We therefore expand the integrand to first order in $\nu^{t}_{\ddot\psi, 0}$ for analytically tractable
expressions. Again, the explicit solutions are not provided for lack of space, however by 
setting the accelerations $a_{g, 0}$ and $a^{ego}_0$ as well as the process noises $\nu^{t}_{\ddot\psi, 0}$, $\nu^{t}_{\dot a, 0}$, $\nu^{ego}_{\ddot\psi, 0}$, and $\nu^{t}_{\dot a, 0}$ to zero, the 
state update expressions as in \cite{Maehlisch_et_al_07} are recovered.



\section{NUMERICAL RESULTS}

\subsection{Simulation setup}

\begin{figure}[t]
\centering
\includegraphics[width=8cm]{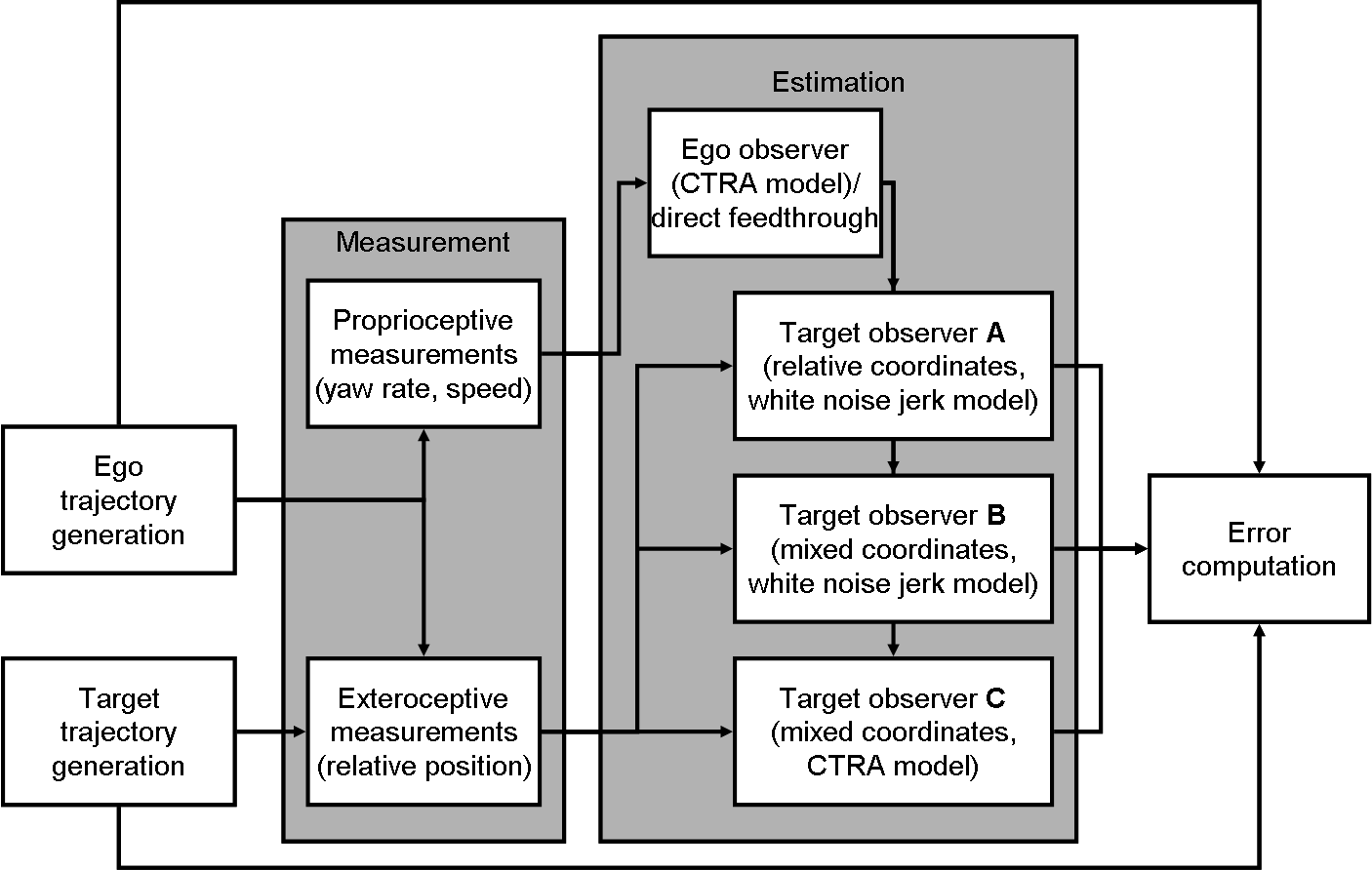}
\caption{Block diagram of the simulation setup.}
\label{fig_simulation_environment}
\vspace{-0.2cm}
\end{figure}

In order to assess the estimation accuracy of the above dynamical models in different coordinate systems, 
a numerical study was performed. First, 50 trajectories with a duration of $20s$ for the ego 
as well as the target vehicle were generated using the CTRA model described in app. \ref{app_CTRA_model}.
The process noise for both CTRA models was chosen to be 
$cov(\nu_{\ddot\psi,k}, \nu_{\dot a,k}) = \diag\left(1\left({rad\over s^2}\right)^2,25\left({m\over s^3}\right)^2\right)$.
In order to perturb the reference trajectories away from the CTRA model, noise was added to the $\psi$ component.
From these
reference trajectories, proprioceptive and exteroceptive measurements corrupted with additive white noise are extracted every $\Delta t_k = 40ms$ - a
typical value for radar or laser measurements. Then the exteroceptive measurements are fed into three extended Kalman 
filters using the
three discrete dynamics introduced in section \ref{sec_examples}, see fig. \ref{fig_simulation_environment}. The contribution of the process noise from the ego dynamics $cov(\nu^{ego}_{\ddot\psi,k}, \nu^{ego}_{\dot a,k})$ as well as
of the relative target dynamics using the CTRA in model C $cov(\nu^t_{\ddot\psi,k}, \nu^t_{\dot a,k})$ were set to the above values for the reference trajectory generation.
The contribution of the relative target dynamics using the white noise jerk dynamics in models A and B were determined by numerically computing
the values of $cov(\nu^t_{\dddot x,k}, \nu^t_{\dddot y,k}) \approx \diag\left(325\left({m\over s^3}\right)^2,325\left({m\over s^3}\right)^2\right)$ over all 50 reference trajectories.

For track initialization, the not-measured entries of the state vector were set to zero.
The proprioceptive measurements are fed into an EKF using the CTRA model for 
the estimation of the ego trajectory. The output of the
proprioception serves as an input or control vector for the exteroceptive observers.
Finally the Euclidean error of the position estimation with respect to the reference
trajectories is determined.

\subsection{Comparison of estimation errors}

At every time step $k$ of every trajectory $j$, the Euclidean error of the estimated relative position
\eq{
\Delta_{jk} = \sqrt{ 
              \left\lVert \begin{pmatrix} x^t_{rel} \cr y^t_{rel} \end{pmatrix} 
             - r \cdot \begin{pmatrix} x^t_{g} - x^{ego}_g \cr y^t_{g} - y^{ego}_g \end{pmatrix} \right\rVert }\nn
}
is computed where $r$ is the rotation matrix as in eq. \Ref{eq_2d_rotation}. 
The estimation performance is then subsumed into the average over all trajectories of the maximal error 
of an individual trajectory - ${\rm av}_j\left(\max_k\left(\Delta_{jk}\right)\right)$ - as well as the average over all trajectories of the 
mean error of an individual trajectory: ${\rm av}_j\left({\rm av}_k\left(\Delta_{jk}\right)\right)$.
\begin{table}[t]
		\begin{tabular}[t]{|c||c|c|c|}
		\hline
		Model & ${\rm av}_j(\max_k(\Delta_{jk}))$ & ${\rm av}_j({\rm av}_k(\Delta_{jk}))$ & $|{\mathcal G}|$\\
		\hline \hline
		A & 7.60 & 1.52 & $1.6\cdot 10^{-11}$ \\
		B & 4.35 & 1.02 & $1.6\cdot 10^{-11}$ \\
		C & 4.17 & 0.95 & $0\dots 3.2\cdot 10^{-3}$ \\
		\hline
		\end{tabular}
	\caption{\rm Estimation errors averaged over all 50 reference trajectories for models A, B, and C.
             Also shown are the values of the determinant of the stochastic observability Gramian ${\mathcal G} = Q^\top \diag( W, \dots, W  )^{-1} Q$ where 
$Q$ is the observability matrix and $W$ the measurement covariance matrix.}
	\label{tab_estimation_errors}
\vspace{-0.7cm}
\end{table}
As can be seen in the first three columns of table~\ref{tab_estimation_errors}, the errors become smaller from model A to model C. Going from purely relative to partially 
relative (``mixed") coordinates (A $\rightarrow$ B) has a larger effect than going from the white noise jerk model for the target dynamics
to the CTRA model in mixed coordinates (B $\rightarrow$ C).

\subsection{Observability analysis} \label{sec_numerical_observability_analysis}

In the introduction it was stated that the target state in global coordinates is unobservable if the ego vehicle has
no absolute position measurements by e. g. GPS and that the target state is poorly observable if GPS-only (without differential corrections)
measurements are available. 

On the other hand, by formulating the target dynamics in a relative coordinate system as in section \ref{sec_relative_vehicle_dynamics}, 
the target dynamics might become observable
if at least relative 2d position measurements are provided by the exteroceptive sensors. 
As can be seen in section \ref{sec_examples}, only
the observable quantities $v_0^{ego}$, $a_0^{ego}$, and $\dot\psi_0^{ego}$ enter the vector fields in eqs. \ref{eq_pure_relative_highest_derivative}, 
\ref{eq_mixed_white_noise_jerk_vector_field}, and \ref{eq_CTRA_mixed_vector_field}.\footnote{As already stated, the dependence of the process noise covariance matrix on $\psi_0^{ego}$
disappears if the Cartesian process noise is chosen to be isotropic.} Hence also the induced norm of the combined input and process noise is bounded -- another
necessary condition for the convergence of the EKF \cite{SongGrizzle95}.

For the observability analysis we have evaluated the determinant of the observability Gramian (see e. g. \cite{Bay99}) over all trajectories.
The determinants for models 
A and B are exactly $960400 (\Delta t_k)^{12}$; this is also the expression for the simple 2d white noise jerk model as in \cite{MatzkaAltendorfer08}.
The determinant for model C turned out to be too complex to be derived analytically and is given by a numerical
range in table~\ref{tab_estimation_errors}.

As can be seen in the last column of table \ref{tab_estimation_errors} models A and B are always observable irrespective of the state, 
whereas model C can become unobservable. This happens for example when speed and acceleration over ground become zero.
However despite the intermittent ill-observability, model C performs best in terms of the estimation error (table \ref{tab_estimation_errors}). 

\section{CONCLUSIONS AND OUTLOOK}

In this paper a general framework for the derivation of the dynamical vector field
of the relative target dynamics for target tracking based on the vector fields for the global target and ego dynamics was 
presented. It was also shown how non-inertial contributions are taken into account when relative (body-fixed)
velocities and accelerations are part of the state vector.
The framework was applied to three different combinations of target and ego dynamics and 
coordinate choices and their ability in tracking targets was assessed by a numerical study.
Model C, the CTRA model formulated in mixed coordinates, i. e. with relative position and angle coordinates
but global (over ground) angle rate, speed, and acceleration coordinates, turned out to be more accurate than
the other two models.
 
Since model C can become unobservable, its observability should be studied more carefully. Although a point in state space where 
the system becomes unobservable has already been identified, an exhaustive characterization of the unobservable
state and input subspace is required along with an analysis how those unobservable subspaces affect
the overall tracking performance in realistic driving scenarios.


\appendix

\subsection{2D rotations and non-inertial contributions} \label{app_rotations}

The 2D rotation matrix is given by
\eq{
r = \begin{pmatrix} \cos(\psi^{ego}) & \sin(\psi^{ego}) \cr
                   -\sin(\psi^{ego}) & \cos(\psi^{ego})  \end{pmatrix} \label{eq_2d_rotation}
}

\begin{figure}[ht]
\centering
\includegraphics[width=8cm]{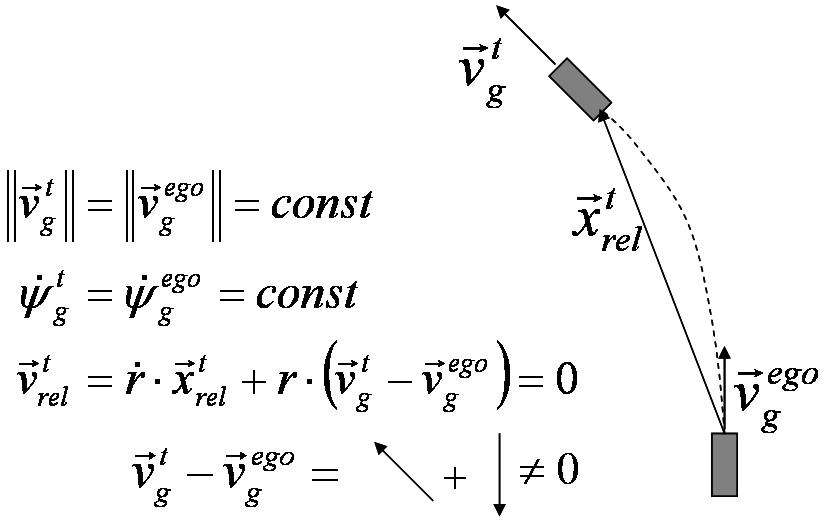}
\caption{ACC target in follow mode for a circular trajectory
with constant yaw rates and constant speeds.}
\label{fig_noninertial_correction}
\vspace{-0.2cm}
\end{figure}

The transformation of a six-dimensional vector containing the differences between target and ego vehicle 
in horizontal position, velocity, and acceleration over ground
into body fixed coordinates is accomplished by the matrix
\eq{
M = \begin{pmatrix} r & 0 & 0 \cr
               \dot r & r & 0 \cr
        \ddot r & 2\dot r & r \end{pmatrix}
}
On the diagonal are the rotation matrices \Ref{eq_2d_rotation} while the off-diagonal terms are due
to non-inertial corrections: e. g. the entries $\ddot r$ and $2\dot r$ give rise to centrifugal and Coriolis
pseudo-forces, respectively, see e. g. \cite{goldstein}. The $\dot r$-term provides a necessary velocity correction
as can be seen in fig. \ref{fig_noninertial_correction}.

Without the correction term which can also be written in 3d vector notation 
as $-\vec{\dot\psi}\times\vec x^t_{rel}$ the relative velocity would not be zero.

\subsection{Constant turn rate and acceleration model} \label{app_CTRA_model}

A special case of a 2d curvilinear motion model \cite{LiJilkov03} is here referred to as the constant turn rate and acceleration model (CTRA);
its simpler version, the constant turn rate and velocity model (CTRV) represents the non-holonomic system of a vertical disk rolling
on a horizontal plane \cite{goldstein}. Both models are commonly used to approximate a wheeled vehicle's dynamics under normal driving conditions
where the slip angle can be neglected \cite{Schubert_et_al_08}. The longitudinal acceleration is included in the state vector
$\xi^{ego}_g = (x,\ y,\ \psi,\ \dot\psi,\ v,\ a)^\top$ since e. g.
ACC systems use the acceleration of the target vehicle as a control input. 
Its continuous dynamics is
given by
\eq{
{d\over dt}\xi^{ego}_g = 
\left(v\cos\psi,\ v\sin\psi,\ \dot\psi,\ \nu_{\ddot\psi}(t),\ a,\ \nu_{\dot a}(t)\right)^\top \label{eq_diff_eq_ego}
}
The discrete-time counterpart solution
\eq{ 
\xi^{ego}_g(t) = F^{ego}_g( \xi^{ego}_g(t_0), t-t_0, \nu^{ego}_g(t_0) ) \label{eq_solution_CTRA}
}
with $\nu^{ego}_g(t) = (\nu_{\ddot\psi}(t),\ \nu_{\dot a}(t))^\top$
can be obtained by linear system techniques and is not provided here for
lack of space. 
It contains Fresnel integrals which can be expanded
to first order in $\nu_{\ddot\psi}(t_0)$ to obtain analytically tractable expressions. The solution also
has inessential singularities at $\dot\psi(t_0) = 0$ which can be removed by Taylor expansion.



\bibliographystyle{IEEEtran}
\bibliography{bibliography}

\end{document}